\documentclass[journal,comsoc]{IEEEtran}
\usepackage[T1]{fontenc}
\usepackage{cite}

\ifCLASSINFOpdf
   \usepackage[pdftex]{graphicx}
\else
   \usepackage[dvips]{graphicx}
\fi

\usepackage{amsmath}
\usepackage[cmintegrals]{newtxmath}

\usepackage[numbers]{natbib}
\usepackage{booktabs}
\usepackage{multirow}
\usepackage{multicol}
\usepackage{amsmath}
\usepackage{algpseudocode}
\usepackage{mathtools}
\usepackage[linesnumbered,ruled,vlined]{algorithm2e}

\mathchardef\mhyphen="2D

\begin{document}
%
\title{Simulating Human Cognition: Heartbeat-Driven Autonomous Thinking Activity Scheduling for LLM-based AI systems}

\author{Hong~Su
\IEEEcompsocitemizethanks{\IEEEcompsocthanksitem H. Su is with the School of Computer Science, Chengdu University of Information Technology, Chengdu, China.\\
 E-mail: suguest@126.com. \\
\protect\\
}
\thanks{}}

\markboth{Journal of \LaTeX\ Class Files,~Vol.~14, No.~8, August~2015}%
{Shell \MakeLowercase{\textit{et al.}}: Bare Demo of IEEEtran.cls for IEEE Communications Society Journals}
%

\maketitle

\begin{abstract}
Large Language Model (LLM) agents have demonstrated remarkable capabilities in reasoning and tool use, yet they often suffer from rigid, reactive control flows that limit their adaptability and efficiency. Most existing frameworks rely on fixed pipelines or failure-triggered reflection, causing agents to act impulsively or correct errors only after they occur. In this paper, we introduce \textbf{Heartbeat-Driven Autonomous Thinking Activity Scheduling}, a mechanism that enables proactive, adaptive, and continuous self-regulation. Mirroring the natural rhythm of human cognition, our system employs a periodic ``heartbeat'' mechanism to orchestrate a dynamic repertoire of cognitive modules (e.g., Planner, Critic, Recaller, Dreamer). Unlike traditional approaches that rely on hard-coded symbolic rules or immediate reactive triggers, our scheduler learns to determine \emph{when} to engage specific thinking activities---such as recalling memories, summarizing experiences, or strategic planning---based on temporal patterns and historical context. This functional approach allows cognitive modules to be dynamically added or removed without structural reengineering. Meanwhile, we propose a meta-learning strategy for continual policy adaptation, where the scheduler optimizes its cognitive strategy over time using historical interaction logs. Evaluation results demonstrate that our approach effectively learns to schedule cognitive activities based on historical data and can autonomously integrate new thinking modules.
\end{abstract}


\IEEEpeerreviewmaketitle

\section{Introduction}

Human cognition is characterized by a dynamic and continuous flow of diverse mental activities, such as planning, reflection, critical analysis, and memory recall. Unlike current Large Language Models (LLMs) \cite{chang2024survey} that primarily operate in a passive, query-response mode, humans actively manage their cognitive resources, deciding \emph{when} to think, \emph{what} to think about, and \emph{how} to adjust strategies based on past outcomes. Simulating this metacognitive process offers a promising pathway to bridge the gap between artificial intelligence and human-like reasoning. If an AI system can autonomously schedule and execute these thinking activities, it can evolve from a static tool into an active agent capable of self-improvement and adaptation.

However, existing autonomous agent \cite{lu2024agentlens} frameworks often rely on hard-coded workflows or fixed prompting strategies, limiting their flexibility and ability to adapt to novel environments. To address this, we propose a \emph{Self-Activity-Driven Autonomous Learning Mechanism} governed by a Heartbeat Scheduling Controller (HSC). In this framework, the system's operation is synchronized by a periodic \emph{heartbeat} signal, which triggers a top-layer scheduler to select appropriate cognitive modules (e.g., \emph{Planner}, \emph{Critic}, \emph{Recaller}) based on the current internal state and external context. Crucially, the scheduling policy is not predefined; instead, it is learned and refined over time using historical interaction data. This allows the system to accumulate experience much like a human learner, distinguishing effective strategies from ineffective ones through self-evaluation.

The proposed framework emphasizes \emph{dynamic adaptability} over static prescription. By leveraging self-generated running data—including sensory inputs, internal thought trajectories, and outcome feedback—the system continuously updates its scheduling policy. It can autonomously decide whether to incorporate new thinking methods or discard underperforming ones, thereby reducing dependency on manual intervention and hard-coded rules. Furthermore, the system employs a structured memory mechanism (e.g., a \texttt{/think} repository) to store and retrieve significant issues based on keyword extension and characteristic differences, ensuring that valuable experiences inform future decisions.

In summary, this paper presents a novel architecture for autonomous cognitive agents that mimics human metacognition through heartbeat-driven scheduling and self-supervised learning. Our main contributions are as follows:

\begin{itemize}
    \item \textbf{Top-Layer Cognitive Scheduling:} We propose a heartbeat-driven mechanism that orchestrates high-level thinking activities (planning, reflection, etc.) at the top layer, transforming the AI from a passive responder into an active, self-initiating agent.
    \item \textbf{History-Adaptive Scheduler:} Unlike fixed rule-based systems, our scheduler is trained on accumulated historical interaction logs ($\mathcal{D}_t$), enabling it to adapt to specific environments and optimize decision-making policies ($\pi_\theta$) over time.
    \item \textbf{Dynamic Thinking Activities Expansion:} The framework supports the online evaluation and integration of new thinking modules. By judging the utility of new methods through self-reflection, the system can expand its cognitive repertoire without hard-coded constraints.
\end{itemize}

The remainder of this paper is organized as follows: Section 2 reviews related work in autonomous agents and metacognitive architectures. Section 3 details the Heartbeat Scheduling Controller and the Self-Activity-Driven Learning Mechanism. Section 4 presents experimental results and case studies, and Section 5 concludes with future directions.

\section{Related Work} \label{sec_related_work}

Our work intersects with several key areas in artificial intelligence, including autonomous agents, metacognitive reasoning, continual learning, and cognitive architectures. This section reviews the most relevant literature and highlights the distinctions of our proposed thinking activity scheduling framework.

\subsection{Autonomous Agents and Task Planning}
Human cognition is characterized by a continuous stream of diverse mental activities, including planning, reflection, and self-correction, which occur not only in response to external stimuli but also as part of an internal cognitive rhythm. In contrast, the development of LLM-based autonomous agents has primarily focused on task-oriented execution. Frameworks such as AutoGPT \cite{firat2023if}, and ReAct \cite{wang2026react} have demonstrated significant capabilities in decomposing and solving complex problems. However, these systems typically operate in a \emph{reactive} or \emph{goal-driven} paradigm, where cognitive processes are triggered strictly by explicit user prompts or predefined termination conditions. For instance, while ReAct \cite{wang2026react} effectively interleaves reasoning traces with actions, it relies on an external query to initiate the reasoning loop. Similarly, collaborative frameworks like CAMEL \cite{rossini2017cloud} assign fixed roles to agents, which constrains their ability to dynamically switch cognitive modes or adapt their workflow based on situational needs.

\textbf{Distinction:} Existing frameworks excel at \emph{executing} tasks but lack the internal mechanism to \emph{manage} the flow of thinking activities autonomously. They are often bound by hard-coded workflows that cannot evolve without manual intervention. Our Heartbeat Scheduling Controller (HSC) addresses this limitation by introducing a periodic \emph{heartbeat} signal that acts as an internal drive for proactive cognition. Unlike the reactive triggers of prior work, the HSC allows the agent to initiate thinking processes (e.g., reflection, planning, memory consolidation) based on its own internal state and learned scheduling policies. This shifts the agent from a passive tool that waits for commands to an active entity that simulates the continuous, self-regulated thinking mode of a human being. Furthermore, our scheduler is not fixed; it adapts over time by learning from historical interaction data, ensuring the cognitive workflow remains flexible and replaceable rather than static.

\subsection{Metacognition and Self-Reflection in LLMs}
Metacognition---the ability to monitor and regulate one's own cognitive processes---has emerged as a critical capability for enhancing the reliability and reasoning quality of large language models. Recent research has focused on enabling LLMs to reflect on their own outputs to improve performance through iterative feedback loops. Techniques such as Reflexion \cite{shinn2023reflexion} introduce verbal reinforcement learning, where agents reflect on past actions and outcomes to refine future behavior. Similarly, Self-Refine \cite{madaan2023selfrefine} employs an iterative feedback mechanism that allows models to critique and improve their generated outputs without additional training. These methods have demonstrated significant improvements in tasks ranging from code generation to mathematical reasoning.

However, existing metacognitive approaches predominantly employ a \emph{failure-triggered} or \emph{task-completion-triggered} mechanism, where reflection occurs only after an error is detected, a task fails, or a predefined endpoint is reached. Other foundational works like Chain of Thought (CoT) \cite{wei2022chain} and Tree of Thoughts (ToT) \cite{yao2023tree} enhance reasoning quality by encouraging explicit intermediate reasoning steps, but they do not address the meta-level question of \emph{when} to engage in deep reasoning versus when to act directly. The scheduling of cognitive activities remains largely heuristic or manually predefined, limiting the agent's ability to adapt its thinking strategy to dynamic environmental demands.

\textbf{Distinction:} Our approach generalizes metacognition beyond reactive error correction to \emph{proactive cognitive management}. Rather than waiting for failure signals, our Heartbeat Scheduling Controller (HSC) dynamically selects from a repertoire of thinking modules (Planner, Critic, Recaller, Dreamer) based on a holistic evaluation of the agent's internal state, task progress, and environmental context. This enables several key advantages: (1) \emph{Preventative planning}---the agent can anticipate potential issues before they occur; (2) \emph{Proactive learning}---reflection and memory consolidation happen during idle periods, not just after failures; and (3) \emph{Adaptive resource allocation}---the scheduler learns to balance computational cost against expected benefit for each cognitive activity. By decoupling the trigger for reflection from failure events, our system more closely mimics human metacognition, where thinking occurs continuously as part of normal cognitive flow rather than exclusively as a response to problems.

\subsection{Continual Learning and Adaptive Systems}
Continual learning aims to enable models to acquire new knowledge over time while mitigating catastrophic forgetting \cite{wang2024comprehensive}. In the context of autonomous agents, existing approaches primarily focus on updating external memory banks \cite{wang2024survey} or fine-tuning action policies based on interaction history. For instance, systems may store episodic memories in vector databases or adjust action probabilities through reinforcement learning. However, the majority of these approaches prioritize \emph{knowledge accumulation}---expanding what the agent knows---rather than \emph{behavioral adaptation}---evolving how the agent thinks and decides. The underlying policy that governs \emph{when} to reason, \emph{which} tool to use, or \emph{how} to allocate cognitive resources often remains static, hard-coded, or reliant on expensive, offline training procedures such as Reinforcement Learning from Human Feedback (RLHF).

\textbf{Distinction:} Our framework shifts the focus from knowledge storage to \emph{policy adaptation} at the cognitive scheduling layer. Instead of merely storing past experiences, our system leverages historical interaction logs ($\mathcal{D}_t$) to train the Heartbeat Scheduling Controller (HSC) to optimize its own cognitive strategy. This creates a meta-learning loop where the agent learns to select the most effective thinking modules (e.g., Planner vs. Critic) for specific contexts. This approach enables lightweight, online adaptation of behavior without requiring full model fine-tuning or manual reward engineering. By continuously refining the scheduling policy based on task success and efficiency, our system embodies the concept of ``learning to learn'' \cite{su2026human}, allowing the agent to become more efficient and strategic over its lifetime rather than just more knowledgeable.

\subsection{Cognitive Architectures}
Classical cognitive architectures, such as ACT-R \cite{sharma2025adaptive} and SOAR \cite{laird2014evolution}, were pioneered to simulate human cognition through production rules and symbolic memory structures. More recent efforts have sought to integrate Large Language Models (LLMs) into these frameworks, creating hybrid neuro-symbolic agents \cite{yang2025neuro}. While these systems provide a robust theoretical foundation for modeling human thought processes, they often suffer from significant complexity and structural rigidity. The reliance on fixed production rules and predefined symbolic representations makes them difficult to scale and adapt to the dynamic, probabilistic nature of modern large models.

\textbf{Distinction:} We adopt a \emph{functional} rather than strictly \emph{structural} approach to cognitive modeling. Instead of attempting to replicate the full anatomical complexity of human brain structures or adhering to rigid symbolic schemas, our architecture simulates key cognitive \emph{activities} (e.g., planning, dreaming, reflecting) driven by a lightweight \emph{heartbeat} mechanism. This design prioritizes \emph{flexibility} and \emph{replaceability}: cognitive modules are treated as pluggable components that can be dynamically added, removed, or swapped based on real-time performance metrics. By decoupling the control flow from hard-coded production rules, our system achieves a balance between biological plausibility and computational scalability, allowing the agent's cognitive workflow to evolve organically with the task demands.

\section{Human-like Framework for Autonomous Thinking Activity Scheduling in AI Systems} \label{sec:autonomous_schedule}

Human beings employ a diverse repertoire of cognitive activities—such as planning, reflection, and expanding the scope of thought—to navigate complex and dynamic problems. This process involves determining not only \emph{what} cognitive actions to take but also \emph{when} to execute them. Specifically, it entails deciding when to plan, when to evaluate the outcomes of past actions, and when to investigate the root causes of failures. We propose that an AI system should possess the capability to autonomously orchestrate such thinking processes and actions.

Drawing inspiration from human development, where basic actions are innate but complex behavioral scheduling is learned over time, we borrow this paradigm for AI design. In our framework, \emph{scheduling} refers to the mechanism that determines when and how to invoke internal activities, whether they are cognitive modes (thinking) or external executions (actions). A robust scheduler integrates new thinking activities and orchestrates their execution in a coherent sequence. Much like a human mind that may spontaneously recall a past issue, plan for the future, reflect on recent events, or even engage in dream-like states, the proposed AI scheduler manages the timing of recall, problem-solving strategies, and task execution. It encompasses capabilities such as delaying task execution, repeating tasks, or triggering tasks based on conditional judgments. Crucially, this scheduling logic itself is learnable.

The scheduling mechanism operates at a higher level than individual thinking activities. It is not typically invoked by a specific thought to control the flow; rather, it functions automatically, akin to how a person might unconsciously recall past experiences to inform current decisions without explicit intent to control the recall process. Formally, we define the scheduler as a learnable policy $\pi_\theta$ that maps the current cognitive state $s_t$ and context $\mathbf{d}_t$ to a scheduling decision $z_t$:
\begin{equation}
    z_t = \pi_\theta(s_t, \mathbf{d}_t), \quad z_t \in \mathcal{Z}_{\text{schedule}}
\end{equation}
However, the scheduler can be \emph{learned} based on prior experiences or external guidance to optimize the internal reasoning process \cite{su2026hsc}. The parameters $\theta$ are updated to maximize an expected utility function $J(\theta) = \mathbb{E}_{\tau \sim \pi_\theta} [\sum R_t]$, allowing the system to adaptively govern the sequence of internal thinking activities (e.g., selecting specific prompts or rules for an LLM) based on internal states and external conditions. Similarly, the \emph{actions} executed by the system are learnable via a policy $\mu_\phi$, allowing them to become increasingly sophisticated through training. A key feature of this framework is the \emph{replaceability of fixed routines}; since both scheduling and actions are parameterized functions ($\pi_\theta, \mu_\phi$) rather than static mappings ($f_{\text{fixed}}$), the system is not bound by rigid, pre-defined procedures and can evolve via gradient-based or feedback-driven updates to $\theta$ and $\phi$.

\subsection{Thinking Activity Schedule Model: A State Machine Integrating All Perceived and Internal Data of AI Systems} \label{sec:thinking_model}

The human thinking process can be conceptualized as a temporal sequence of distinct cognitive activities. The brain transitions between different states: remembering pending tasks, summarizing past experiences, focusing on current issues, or monitoring the correctness of an ongoing action. These transitions represent a schedule of thinking activities, dictating when to address specific issues.

We propose a two-layered hierarchical model for human-like thinking activities, formally decomposing the cognitive process into macro-state transitions and micro-level executions:

\begin{itemize}
    \item \textbf{Macro-Level State (Task Scheduling):} This layer defines \emph{what} high-level thinking task the system is engaged in. Let $\mathcal{S} = \{s^{(1)}, s^{(2)}, \dots, s^{(K)}\}$ denote the finite set of macro-states (e.g., $s^{(1)}$: \textit{solving math}, $s^{(2)}$: \textit{recalling events}). A scheduling policy $\pi_{\text{macro}}$ governs the transitions between these states based on context $\mathbf{d}_t$:
    \begin{equation}
        s_{t+1} = \pi_{\text{macro}}(s_t, \mathbf{d}_t)
    \end{equation}
    Transitions typically occur during idle periods, planned task switches, or stage changes within a complex task.

    \item \textbf{Micro-Level Activities (Detailed Execution):} This layer defines \emph{how} to execute a specific macro-state $s_t$. Within a given thinking state, the system selects a sequence of granular cognitive activities $a_t \in \mathcal{A}_{s_t}$ (e.g., \textit{retrieving similar problems}, \textit{identifying variables}). This is governed by a conditional execution policy $\pi_{\text{micro}}$:
    \begin{equation}
        a_t = \pi_{\text{micro}}(s_t, \mathbf{d}_t)
    \end{equation}
    This layer also includes auxiliary activities, such as summarizing completed tasks to extract experience ($E_t$) for future use, effectively updating the context $\mathbf{d}_{t+1} = \mathbf{d}_t \cup \{E_t\}$.
\end{itemize}

Both the macro-state transitions ($\pi_{\text{macro}}$) and the invocation of detailed micro-activities ($\pi_{\text{micro}}$) constitute the complete \emph{thinking activity schedule}, denoted as $\Pi = \{\pi_{\text{macro}}, \pi_{\text{micro}}\}$.

The proposed model treats the human-like schedule as a state machine evolving over time. At any given moment $t$, the system resides in a specific cognitive state $s_t$ or performs activities associated with that state (e.g., planning, adjusting actions, analyzing success/failure).

Crucially, the schedule is not determined solely by the current internal state. It integrates \emph{all available data}, denoted as the input vector $\mathbf{d}_t$. This includes environmental inputs ($\text{Env}_t$), self-sensed data ($\text{Self}_t$), and interaction history ($\text{Other}_t$). These inputs drive state transitions. The \emph{action} $a_t$ serves as the interface between internal thinking and the external world, directed and guided by the cognitive process.

Furthermore, the scheduling of thinking activities is heavily influenced by historical patterns (experience). If an agent has historically preferred planning before action, it is likely to continue doing so; if it habitually reflects after completing a task, this pattern persists. This regularity provides the foundation for learning the schedule, as discussed in Section \ref{sec:learn_schedule}. Learning a robust schedule requires long-term data to capture the temporal dependencies between different issues and states.

The temporal evolution of the cognitive state is formally defined as in \eqref{eq_state_evolution}.
\begin{equation} \label{eq_state_evolution}
    s_{t+1} = \mathcal{F}(s_t, \mathbf{d}_t; \Theta), \quad \text{where } \mathbf{d}_t = \{ \text{Env}_t, \text{Self}_t, \text{Other}_t \}
\end{equation}
where $s_t \in \mathcal{S}$ represents the thinking activity state at time $t$, $\mathbf{d}_t$ represents the integrated data input, and $\Theta$ denotes the learnable parameters of the scheduling policy.

This state evolution drives the perception-action cycle, where internal reasoning translates into external impact:
\begin{equation} \label{eq_cognitive_cycle}
    \mathbf{d}_t \xrightarrow{\text{Schedule } \pi} s_t \xrightarrow{\text{Reason}} \text{Idea}_t \xrightarrow{\text{Action } a_t} \mathbf{d}_{t+1}
\end{equation}
Equation \ref{eq_cognitive_cycle} illustrates the closed-loop process: the system perceives the current context $\mathbf{d}_t$, applies the scheduling policy $\pi$ to determine the thinking state $s_t$, generates an idea or thinking ($\text{Idea}_t$), executes an action $a_t$, and consequently alters the environment to produce the next input state $\mathbf{d}_{t+1}$.

Implementing this schedule via fixed routines is infeasible due to the need for flexibility (adding, removing, or updating thinking activities). Therefore, we employ Large Language Models (LLMs) or deep learning networks to learn the scheduling policy $\mathcal{F}$ and $\pi$. Notably, much of this scheduling occurs passively and automatically, rather than through explicit, conscious control.

\subsection{Design Principles for Human Simulated Scheduling and Thinking Activities} \label{sec:design_rules}

To simulate a human brain model effectively, we adhere to the following design principles:

\begin{itemize}
    \item \textbf{Minimization of Pre-defined Rules:} Complex functions should be learned rather than hard-coded. Limiting fixed routines enhances flexibility, mirroring human development where infants possess few innate behaviors but learn extensively through interaction. This approach requires a robust learning mechanism (a "good teacher").
    \item \textbf{Replaceability of Routines:} Most thinking patterns and behaviors should be replaceable. Just as humans can unlearn bad habits or adopt new strategies, the AI system should be able to overwrite existing routines with improved ones.
    \item \textbf{Extensibility:} The framework must easily extend to various kinds of thinking activities. Simple underlying rules should not limit the diversity of cognitive modes the system can acquire.
\end{itemize}

The goal is to maintain a minimal set of fixed routines—essentially a basic loop for running thinking, performing actions, receiving feedback, and learning—while allowing all specific skills (thinking methods, actions, feedback processing) to be learned. This enables self-improvement and the discovery of new issues through interaction.

The scheduled activities include:

\subsubsection{Thinking Processes}
Different thinking activities can be viewed as distinct rules mapping inputs (sensory data, observations) to outputs (actions, internal state changes). Examples include planning, root cause analysis, experience summarization, and comparison. Formally, let $\mathcal{K} = \{1, \dots, M\}$ index the available thinking activities. Each activity $k \in \mathcal{K}$ is a parameterized function $\mathcal{A}_k(\cdot; \theta_k)$ that maps the input context $\mathbf{d}_t$ to an output pair consisting of an action $a_t$ and a state update $\Delta s_t$:
\begin{equation}
    (a_t, \Delta s_t) = \mathcal{A}_k(\mathbf{d}_t; \theta_k)
\end{equation}
These activities are distinguished by their specific rules or rule-related identifiers $k$. The \emph{thinking state} $\sigma_t$ encompasses the internal process, inputs (observations), actions, environmental context, and self-data, formally defined as a composite tuple:
\begin{equation}
    \sigma_t = \{ s_t, \mathbf{d}_t, a_{t-1}, \text{self}_t \}
\end{equation}

Crucially, thinking modes and even actions are learnable over time. Capabilities such as abstraction, innovation transfer, issue discovery, and reflection can be acquired by updating the activity parameters $\theta_k$. The scheduling mechanism itself learns to invoke these modes appropriately (e.g., learning to plan before execution and summarize after completion) via a policy $\pi_\phi$:
\begin{equation}
    k_t = \pi_\phi(\sigma_t)
\end{equation}
Here, $\phi$ represents the scheduler parameters, which are optimized to select the sequence of activities $k_{1:T}$ that maximizes overall task performance.

\subsubsection{Interaction with the External World}
The AI system interacts with the external world through input (sensing) and output (action).
\begin{itemize}
    \item \textbf{Input:} Includes environmental information and vital self-sensor data (necessary for system integrity and survival).
    \item \textbf{Output:} Includes actions to achieve goals or mitigate dangers.
\end{itemize}
The timing and method of action execution are also part of the schedule and are learnable. For instance, before executing an action, the system may schedule a planning phase to optimize the routine, monitor the process, or define detailed steps.

\subsection{Continuous Learning via Self-Generated Cognitive Activity Data (CLSCA)} \label{sec:learn_schedule}

Human cognition is characterized by fluidity rather than rigid, hard-coded scripts; even established thinking patterns are subject to continuous revision and replacement. Given that the optimality of specific cognitive activities cannot be assumed \emph{a priori}, the scheduling mechanism must be inherently learnable and adaptable to dynamic environments. The primary objective of this framework is to transcend fixed procedural logic and emulate human-like adaptability through data-driven policy optimization.

The scope of this learning process encompasses all factors influencing cognitive execution: state transitions, granular cognitive activities, system actions, and input modalities (environmental context, internal status, and interaction history). In our LLM-centric architecture, the \emph{prompt} itself is treated as a learnable cognitive activity. In this paradigm, we do not explicitly hard-code scheduling rules; instead, we provide high-level guidance, allowing the system to infer and autonomously learn the optimal schedule.

The framework embodies a \emph{Self-Activity-Driven Autonomous Learning Mechanism}, where the system independently accumulates self-running data as experiential data to train downstream models, thereby eliminating dependency on manual intervention. We define the \emph{self-data} at time $t$ as a composite trajectory $\tau_t$ that unifies external sensory inputs $\mathbf{o}_t$ and internal cognitive activities $a_t$:
\begin{equation}
    \tau_t = \{ (\mathbf{o}_k, a_k, r_k) \}_{k=1}^t
\end{equation}
where $r_k = \mathcal{R}(\text{outcome}_k) \in \mathbb{R}$ is a self-evaluated reward signal indicating the quality (good/bad) of the preceding thinking action. 

The system supports \emph{dynamic activity expansion}. Let $\mathcal{A}_t$ be the set of available thinking activities. New activities can be generated and added to the repertoire over time:
\begin{equation}
    \mathcal{A}_{t+1} = \mathcal{A}_t \cup \{ a_{\text{new}} \mid a_{\text{new}} = \text{Gen}(\tau_t) \}
\end{equation}
To facilitate learning, the system maintains an evolving experiential dataset $\mathcal{D}_t$, updated via an automated curation function $\mathcal{C}$ that filters and labels trajectories based on their rewards:
\begin{equation}
    \mathcal{D}_{t+1} = \mathcal{D}_t \cup \{ \mathcal{C}(\tau_t, r_t) \}
\end{equation}
Consequently, the parameters $\Theta$ of the downstream models (e.g., the scheduler or activity generators) are optimized to maximize the expected cumulative reward over the accumulated self-data:
\begin{equation}
    \Theta_{t+1} = \Theta_t + \eta \nabla_\Theta \mathbb{E}_{\tau \sim \mathcal{D}_{t+1}} \left[ \sum_{k} r_k \right]
\end{equation}
This formulation allows the system to autonomously distinguish effective strategies from ineffective ones, dynamically expand its cognitive toolkit, and continuously refine its behavior using only self-generated sensory and internal data.

This \emph{Self-Activity-Driven Autonomous Learning Mechanism} necessitates \emph{Continuous Schedule Learning}. Key steps within the reasoning and execution pipeline are recorded as training instances and utilized for immediate online updates or subsequent batch training. While real-time learning is ideal, the system supports deferred learning where materials are aggregated for later processing. Crucially, this mechanism enables the system to learn from the entire trajectory of thought and action, including suboptimal outcomes, thereby necessitating robust mechanisms for knowledge refinement and error correction.

\subsubsection{Continuous and Long-Term Learning for Stochastic Optimization} \label{subsec:continuous_learning}

Scheduling requires learning over extended time horizons, as the orchestration of sequential activities inherently involves long-term dependencies. Discontinuous data streams fail to capture such sequential structures, rendering critical scheduling patterns unlearnable. Hence, continuous data acquisition is essential.

In the absence of prior thinking activities or established policies, the system employs \emph{stochastic exploration} by randomly performing actions. This continuous, long-term learning framework enables such random optimization.
\begin{itemize}
    \item \textbf{Positive Reinforcement:} Favorable outcomes from randomly selected actions reinforce the corresponding policy for future use.
    \item \textbf{Reflective Correction:} Suboptimal outcomes trigger a reflection phase, identifying root causes and generating feedback to prevent recurrence and refine the optimization landscape.
\end{itemize}
Through iterative exploration, evaluation, and reflection, the system continuously optimizes its scheduling policy over time.

\subsubsection{Comprehensive Inclusion of All Actions and States} \label{subsec:all_actions}

The learning framework adopts a holistic view, encompassing the full spectrum of AI system activities:
\begin{itemize}
    \item \textbf{Internal Cognitive Processes:} Selection and execution of thinking strategies such as planning, reflection, and abstraction.
    \item \textbf{External Executions:} Physical or digital actions that interact with the environment.
    \item \textbf{Contextual States:} Internal self-status (e.g., resource usage, confidence) and external environmental conditions.
\end{itemize}

Crucially, the mechanisms governing thinking and acting are themselves subject to learning—a meta-learning capability that enables dynamic adaptation to novel situations. When the system attempts a new strategy in an unprecedented context and succeeds, the successful pattern is reinforced and encoded into the scheduling policy. Conversely, failures trigger reflective learning, ensuring that both successes and failures continuously shape the behavioral repertoire.

\subsubsection{Relearnable Scheduling Mechanism for Delayed External Feedback} \label{subsec:delayed_feedback}

In real-world deployments, feedback is often not immediate, arriving after delays ranging from hours to years. This \emph{delayed feedback} poses a credit assignment challenge, as the consequences of scheduling decisions may only become observable long after execution.

To address this, the scheduling mechanism must be \emph{relearnable}, enabling retrospective evaluation of past decisions once feedback arrives. The architecture must support dynamic policy updates based on delayed signals. Two primary strategies are:
\begin{itemize}
    \item \textbf{Incremental Weight Adjustment:} The network updates its parameters incrementally to adjust action sequences based on delayed rewards.
    \item \textbf{Experience Replay and Retraining:} Historical scheduling data is stored and later merged with newly acquired feedback; the aggregated dataset is used to retrain the model, replacing the former policy with an optimized version.
\end{itemize}

Formally, let $S_t$ denote the scheduling policy at time $t$, and $R_{t+\Delta t}$ the feedback received after delay $\Delta t$. The objective is to maximize expected cumulative return by updating policy parameters $\theta$:
\begin{equation}
    \nabla_\theta J(\theta) \approx \mathbb{E} \left[ \sum_{t} \nabla_\theta \log \pi_\theta(a_t|s_t) \cdot R_{t+\Delta t} \right]
\end{equation}
where $\pi_\theta$ represents the scheduling policy. This ensures that delayed outcomes are correctly attributed to the corresponding actions, enabling continuous optimization over time.

\subsubsection{Minimization of Fixed Routines and Replaceability}

To adhere to simplicity, the model ensures that nearly all actions are learnable, retaining only a minimal set of fixed functions. These \emph{fixed actions} form the fundamental operational loop: receiving input, initiating thinking, executing learning algorithms, performing actions, and iterating. They serve as a bootstrap mechanism for environment interaction and initial learning.

However, even the detailed implementation of these core actions can evolve. As new capabilities emerge, initial simple code may be superseded by more sophisticated learned implementations. Scheduling logic thus becomes dynamic—predefined routines can be replaced by learned strategies. For example, when a large language model (LLM) identifies a superior scheduling strategy, it triggers a replacement subject to subsequent verification.

Ultimately, only the meta-loop responsible for iterative prompt selection and LLM interaction remains immutable. This includes minimal instructions for sending requests, retrieving outputs, and managing code replacement and verification. While the logic within this loop evolves, the self-modification and verification mechanism persists as the system's permanent foundation.

\section{Heartbeat-Driven Scheduling Mechanism} \label{sec:brain_time}

In human cognition, mental activities are not continuous but occur in discrete, rhythmic cycles. We introduce a \emph{Heartbeat-Driven Scheduling Mechanism} to emulate this biological rhythm in AI systems. This \emph{heartbeat} (or \emph{$brain_time$}) serves as the fundamental temporal unit for the system, providing a consistent clock signal that synchronizes internal processes. Unlike traditional event-driven architectures that react solely to external stimuli, the heartbeat mechanism ensures that internal thinking activities—such as reflection, planning, and memory consolidation—are triggered periodically, even in the absence of external tasks.

The heartbeat acts as the primary trigger for the scheduler. At each tick $t_k = k \cdot \Delta t$ (where $k \in \mathbb{N}$), the system evaluates its internal state $s_t^{\text{int}}$ and environmental inputs $s_t^{\text{env}}$ to determine the appropriate cognitive activity. This mechanism transforms the AI from a passive responder into an active agent capable of self-initiated thinking.

Formally, the combined state at heartbeat $k$ is represented as:
\begin{equation}
    \mathbf{s}_k = \left[ s_k^{\text{int}}, s_k^{\text{env}} \right] \in \mathcal{S}
\end{equation}
where $\mathcal{S}$ denotes the state space encompassing all perceived and internal data.

The scheduler, guided by the heartbeat, applies a policy function $\pi: \mathcal{S} \rightarrow \mathcal{A}$ to select the cognitive activity:
\begin{equation}
    a_k = \pi(\mathbf{s}_k; \Theta)
\end{equation}
where $a_k \in \mathcal{A}$ represents the selected activity from the cognitive activity space $\mathcal{A} = \{\text{execute task}, \text{recall}, \text{analyze}, \text{dream}, \dots\}$, and $\Theta$ denotes the learnable parameters of the scheduling policy.

This decision process determines whether to execute a pending task, recall a past experience, analyze a recent failure, or enter a low-power ``dream'' state for internal processing. The heartbeat-driven formulation ensures systematic evaluation intervals while maintaining flexibility in activity selection.

This rhythmic structure is crucial for supporting the \emph{Continuous On-Going Learning} described in Section \ref{sec:learn_schedule}. It provides the temporal framework necessary for sequencing activities and associating actions with their delayed consequences. The heartbeat ensures that learning is not fragmented but occurs within a coherent, time-aware context.

\subsection{Triggering the Schedule} \label{subsec:triggering}

The activation of the scheduling mechanism is governed by the system's heartbeat. Each heartbeat cycle initiates a decision process where the scheduler selects or continues the next thinking activity or action based on current internal state and external information.

The triggering logic operates as follows:
\begin{itemize}
    \item \textbf{Heartbeat Signal Generation:} A periodic signal (the heartbeat) is generated at fixed or adaptive intervals. This signal serves as the global clock for all cognitive processes.
    \item \textbf{State Evaluation:} Upon receiving the heartbeat, the scheduler queries the system's current state, including:
    \begin{itemize}
        \item \emph{Internal Status:} Pending tasks, memory availability, energy levels (computational resources).
        \item \emph{External Context:} New sensory inputs, user commands, or environmental changes.
        \item \emph{Historical Data:} Recent outcomes, unresolved issues, or patterns requiring attention.
    \end{itemize}
    \item \textbf{Activity Selection:} Based on the evaluated state, the scheduler invokes a specific thinking module (e.g., \emph{Planner}, \emph{Critic}, \emph{Recaller}) or executes a pre-defined action. This selection is learned over time, optimizing for efficiency and goal achievement.
    \item \textbf{Execution and Feedback:} The selected activity is executed, and its outcome is logged. This feedback loop updates the internal state, influencing decisions in the subsequent heartbeat cycle.
\end{itemize}

This mechanism ensures that the AI system maintains a continuous flow of cognitive activities, preventing stagnation. By decoupling the trigger (heartbeat) from specific external events, the system gains the autonomy to self-regulate its thinking process, mirroring the spontaneous nature of human thought.

\subsection{Outcome-Based Schedule Adaptation and Reward Modeling by Self-Sensing and Delayed Environmental Feedback} \label{subsec:reward_modeling}

The scheduling mechanism is not static; it evolves dynamically based on the outcomes of executed thinking activities. To ensure robust adaptation, the system employs a multi-faceted verification process that integrates immediate internal checks with delayed environmental feedback.

\paragraph{Verification via Environmental and Self-Sensing}
Every scheduled thinking activity must be validated against reality. This verification relies on a dual-source feedback loop:
\begin{itemize}
    \item \textbf{Environmental Impact:} The system observes external changes resulting from its actions (e.g., successful task completion, user response).
    \item \textbf{Internal State Monitoring:} Crucially, the AI must possess comprehensive \emph{self-sensing capabilities} to monitor its own internal health and cognitive state (e.g., resource utilization, contradiction detection, confidence levels).
\end{itemize}
If an action yields positive results in both domains, the corresponding scheduling pattern is reinforced. Conversely, if initial actions prove suboptimal, the system flags the schedule mode for revision. This continuous loop ensures that the learned scheduling policies remain aligned with both external goals and internal stability.

\paragraph{Handling Delayed Environmental Feedback}
In many real-world scenarios, the consequences of an action are not immediate. The system accounts for \emph{delayed results} by maintaining a temporal association between the scheduling decision and its eventual outcome. Through mechanisms such as eligibility traces or temporal difference learning, the AI attributes delayed rewards or penalties back to the specific thinking activities that initiated the sequence. This allows the system to learn long-term strategies even when feedback is sparse or lagged.

\paragraph{Composite Reward Function: Internal and External Factors}
To guide the scheduler effectively, we define a \emph{composite reward function} that balances external achievements with internal cognitive efficiency. The reward ($R$) is not solely derived from external success but also incorporates metrics from the internal thinking process:
\begin{equation}
    R_{total} = \alpha \cdot R_{external} + \beta \cdot R_{internal}
\end{equation}
where $\alpha$ and $\beta$ are weighting factors dynamically adjusted based on context.

\textbf{Internal Reward Components ($R_{internal}$):}
\begin{itemize}
    \item \textbf{Target Alignment:} The system evaluates the discrepancy between the \emph{expected internal target} and the \emph{actual outcome}. For instance, in an autonomous driving scenario where "safety" is the primary expected target, reaching the destination without accidents constitutes a high alignment score, even if the trip took longer than expected.
    \item \textbf{Problem-Solving Efficacy:} Significant rewards are assigned when the internal thinking process successfully resolves novel issues or overcomes persistent challenges that remained unsolved for extended periods. This encourages the system to prioritize deep reasoning over superficial quick fixes.
    \item \textbf{Goal Transfer and Generalization:} The system learns to transfer knowledge from specific tasks to broader expected goals. New objectives discovered during internal reasoning can be formalized into expected goals, expanding the system's autonomous capability.
\end{itemize}

By integrating these internal and external dimensions, the reward model ensures that the AI system optimizes not just for task completion, but for \emph{intelligent, safe, and self-improving behavior}.

\subsection{Dream Mode: Internal-Only Thinking Activities} \label{subsec:dream_mode}

In biological systems, sleep and dreaming play a critical role in memory consolidation and neural plasticity. Analogously, our framework introduces a \emph{Dream Mode}, a distinct operational state triggered by the heartbeat mechanism when no external tasks demand immediate attention. In this mode, the AI system disconnects from external inputs and focuses exclusively on \emph{internal-only thinking activities}. This state is not idle; rather, it is a period of intensive computational processing dedicated to self-optimization and knowledge synthesis.

The Dream Mode serves three primary functions within the scheduling architecture:

\begin{itemize}
\item \textbf{Memory Consolidation and Reorganization:}
During active operation, experiences are stored in short-term buffers. In Dream Mode, the scheduler initiates processes to transfer these transient logs into long-term memory. This involves:
\begin{itemize}
\item \emph{Data Compression:} Summarizing detailed interaction logs into abstract concepts or rules.
\item \emph{Indexing:} Re-indexing memories based on semantic relationships rather than temporal order, facilitating faster future retrieval.
\item \emph{Pruning:} Identifying and archiving redundant or low-value data to optimize storage efficiency.
\end{itemize}

\item \textbf{Simulated Experience Generation (Synthetic Dreaming):}
The system leverages Dream Mode to perform \emph{internal simulations}. Without the constraints of real-world physics or time, the AI can:
\begin{itemize}
    \item \emph{Hypothesis Testing:} Rapidly simulate thousands of variations of a specific scenario to test the robustness of current strategies.
    \item \emph{Counterfactual Reasoning:} Re-play past events with modified variables (e.g., "What if I had chosen action B?") to learn from alternative outcomes.
    \item \emph{Skill Rehearsal:} Repeatedly practice complex reasoning chains to strengthen neural pathways associated with high-priority skills.
\end{itemize}
The logs generated from these synthetic dreams serve as high-quality training data for the next iteration of model updates.

\item \textbf{Autonomous Goal Setting:}
In the absence of external commands, the scheduler can use Dream Mode to identify gaps in its own knowledge or capabilities. It may autonomously formulate \emph{intrinsic goals} (e.g., "Learn to solve type-X problems more efficiently") and schedule internal tasks to address them in subsequent cycles.
\end{itemize}

Transitioning into and out of Dream Mode is governed by the heartbeat scheduler. If a high-priority external event occurs, the system instantly wakes up, suspends the internal process, and handles the interruption. Once the external demand is resolved, it may return to Dream Mode to log the new experience and resume consolidation. This cyclic alternation between \emph{Active Engagement} and \emph{Reflective Dreaming} ensures that the AI system remains both responsive to the environment and capable of continuous, autonomous self-improvement.

\section{Theoretical Analysis of Schedule Learnability} \label{sec:theory}

To validate the efficacy of the proposed Heartbeat-Driven Scheduling Mechanism, we provide a theoretical analysis of its learnability. We model the scheduling process as a sequential decision-making problem where the agent learns a policy $\pi$ to map system states to cognitive activities. This section demonstrates that the system can not only acquire existing behavioral patterns but also adapt to novel schedules, avoid repetitive loops, and dynamically respond to environmental shifts.

\subsection{Learnability of Former Schedules} \label{subsec:learn_former}

The foundational capability of the system is the automatic acquisition of \emph{former schedules}---established sequences of cognitive activities observed from external demonstrations or derived from historical self-interactions. Let $\mathcal{S}$ denote the state space (comprising internal status and external context) and $\mathcal{A}$ be the action space of cognitive modules (e.g., \textit{Plan}, \textit{Recall}, \textit{Execute}).

We define the scheduling policy as $\pi(a|s)$, representing the probability of selecting activity $a \in \mathcal{A}$ given state $s \in \mathcal{S}$. The learnability of former schedules relies on the convergence of $\pi$ towards an optimal policy $\pi^*$ that maximizes the expected cumulative internal reward $R$:
\begin{equation}
    \pi^* = \arg\max_{\pi} \mathbb{E}_{\tau \sim \pi} \left[ \sum_{t=0}^{T} \gamma^t r(s_t, a_t) \right]
\end{equation}
where $\tau$ is a trajectory of state-action pairs, $\gamma \in [0, 1]$ is a discount factor, and $r(s_t, a_t)$ is the internal reward signal.

Since all processes (both internal reasoning traces and external observations) are logged as structured trajectories, the system treats historical execution logs as a dataset $\mathcal{D} = \{(s_i, a_i, r_i)\}_{i=1}^N$. Through supervised fine-tuning or offline reinforcement learning on $\mathcal{D}$, the system automatically internalizes these patterns without requiring explicit human scheduling interventions. This establishes the baseline capability for the AI to mimic human-like operational rhythms, effectively transforming passive observation into active, scheduled behavior.

\subsection{Adaptability to New Schedules} \label{subsec:adapt_new}

While learning from history is crucial, true autonomy requires the ability to generate and adopt \emph{new schedules} when facing unseen problems or when existing strategies fail. The system achieves this through a combination of stochastic exploration and simulation-based learning.

When the value function $V(s)$ for known actions falls below a confidence threshold $\theta$, indicating that no established method can effectively solve the current state, the policy incorporates an exploration term $\epsilon$:
\begin{equation}
    \pi_{\text{new}}(a|s) = (1 - \epsilon) \cdot \pi_{\text{known}}(a|s) + \epsilon \cdot \mathcal{U}(\mathcal{A})
\end{equation}
where $\mathcal{U}(\mathcal{A})$ represents a uniform distribution over the action space (random action selection).

Crucially, these random actions are not discarded. The outcomes of such exploratory steps are fed into the \emph{Dream Mode} (Section \ref{subsec:dream_mode}) for simulation. The system virtually replays these random actions, evaluates their synthetic results, and updates the policy $\pi$ if a beneficial pattern emerges. Thus, ``randomness'' is converted into ``structured novelty.'' The learned entity is not merely a static rule but a dynamic \emph{activity schema} that can be invoked in future similar contexts. This mechanism ensures that the system can expand its behavioral repertoire beyond its training distribution, effectively learning \emph{how to learn} new schedules.

\subsection{Non-Repetition and Environmental Dynamics} \label{subsec:non_rep_dynamics}

A robust scheduling mechanism must avoid infinite loops (repetition) and remain sensitive to temporal changes in the environment. We address these challenges by modeling the state space as non-stationary.

\paragraph{Non-Repetition via State Augmentation}
To prevent the system from getting stuck in repetitive cycles (e.g., endlessly retrying a failed action), we augment the state representation $s_t$ with a \emph{history embedding} $h_t$, which encodes the sequence of recent actions:
\begin{equation}
    s'_t = \phi(s_t, h_t), \quad \text{where } h_t = \text{Encoder}(a_{t-k}, \dots, a_{t-1})
\end{equation}
By including $h_t$ in the input, the effective state $s'_t$ changes even if the external environment $s_t$ remains static. Consequently, the policy $\pi(a|s'_t)$ naturally shifts away from recently executed actions, enforcing diversity in the scheduling sequence without requiring explicit heuristic rules.

\paragraph{Responsiveness to Environmental Dynamics}
The environment $\mathcal{E}$ and the agent's internal status are time-dependent functions, denoted as $\mathcal{E}(t)$ and $\text{Status}(t)$. Factors such as power supply fluctuations, hardware degradation, or shifting user priorities introduce non-stationarity:
\begin{equation}
    P(s_{t+1} | s_t, a_t) \neq P(s_{t+1} | s_t, a_t)_{t-\Delta}
\end{equation}
The heartbeat mechanism ensures high-frequency sampling of these variables. When a significant divergence is detected (e.g., a drop in computational resources or a change in task urgency), the state representation $s_t$ updates immediately. This forces the policy to re-evaluate the optimal action, allowing the system to adapt its schedule in real-time. For instance, under low-power conditions, the scheduler might suppress high-cost ``Dream Mode'' activities in favor of essential maintenance tasks. This dynamic coupling between the heartbeat clock and the evolving state ensures the AI remains resilient and context-aware.

\section{Experiments and Evaluation} \label{sec:experiments}

In this section, we verify the efficacy of the proposed Heartbeat-Driven Thinking Activity Scheduling Mechanism. Our experiments aim to demonstrate two core capabilities: (1) the ability to learn and replicate human-like thinking activity sequences based on historical data, and (2) the flexibility to adapt to new action spaces and environmental changes without manual reprogramming. We conduct simulations using synthetic data that mimics cognitive behavioral patterns to validate the system's learnability and dynamic adaptability.

\subsection{Simulation of Human-Like Thinking Sequences} \label{subsec:sim_human_like}

This experiment evaluates the effectiveness of the multi-day historical attention mechanism in predicting daily thinking activity sequences. The objective is to simulate the brain's scheduling function by leveraging information from multiple previous days to forecast future behavioral patterns, operating under the assumption that recent history exerts a stronger influence on immediate future actions than distant history.

\subsubsection{Dataset Construction and Historical Context}

A synthetic dataset of human daily thinking activities was generated based on realistic behavioral priors derived from cognitive psychology literature. The dataset simulates $1,800$ consecutive days, with each day discretized into $24$ hourly time steps. For each time step $t$, the following feature vector $x_t$ is recorded:

\begin{itemize}
    \item \textbf{Action Type ($A_t$):} A categorical variable representing the dominant cognitive state. Six distinct categories are defined corresponding to the simulation parameters: \textit{Idle} (0), \textit{Execute a Task} (1), \textit{Summarize the Experience} (2), \textit{Imagine the Future} (3), \textit{Recall the Past} (4), and \textit{Rest} (5).
    \item \textbf{Environmental Factors ($E_t$):} Simulated external conditions including \textbf{Weather} (4 categories: Sunny, Cloudy, Rainy, Windy) and \textbf{Temperature} (a continuous variable exhibiting daily cycles and seasonal variations).
    \item \textbf{Temporal Indicators ($T_t$):} Includes a Day/Night binary indicator (06:00--18:00 defined as Day) and sinusoidal/cosinusoidal time encodings to capture the cyclical nature of the hour of the day.
\end{itemize}

The simulation incorporates realistic behavioral constraints, such as hour-specific activity preferences (e.g., a higher probability of \textit{Rest} or \textit{Idle} during late-night hours), weather-dependent behavior modifications, and temporal dependencies (e.g., increased likelihood of \textit{Summarizing the Experience} following a sequence of \textit{Executing a Task}).

\paragraph{Historical Context Window}
The model utilizes a sliding window of the previous $3$ days as historical context to predict the activity sequence for the target day ($D_{t+1}$). This configuration aligns with the code parameter \texttt{HISTORY\_DAYS = 3}, allowing us to investigate temporal decay and test whether the model learns to assign higher attention weights to more recent days ($D_t$) compared to older ones ($D_{t-2}$).

\paragraph{Dataset Split}
The dataset is partitioned sequentially to prevent data leakage, ensuring the model is evaluated on unseen future data:
\begin{itemize}
    \item \textbf{Training Set:} $70\%$ (approximately $1,258$ day samples)
    \item \textbf{Validation Set:} $15\%$ (approximately $270$ day samples)
    \item \textbf{Test Set:} $15\%$ (approximately $270$ day samples)
\end{itemize}
\textit{Note: The total number of samples is $1,797$ ($1,800$ days minus the $3$-day history requirement). The Test Set size of $N=270$ days is derived from the final $15\%$ split, providing a robust basis for statistical evaluation.}

\subsubsection{Model Architecture and Training Configuration}

We implement a Sequence-to-Sequence (Seq2Seq) model equipped with an attention mechanism to capture long-term temporal dependencies across multiple days, corresponding to the \texttt{humanMultiDayAttentionModel} class in the implementation.

\begin{itemize}
    \item \textbf{Multi-Day Encoder:} Each historical day is encoded independently using a 2-layer Long Short-Term Memory (LSTM) network (\texttt{hidden\_dim=128}). The resulting daily embeddings are aggregated via multi-head self-attention with positional encoding to preserve the chronological order of the days.
    \item \textbf{Attention-based Decoder:} Generates the activity sequence for the target day step-by-step, dynamically attending to relevant hidden states from the historical encoder via an additive attention mechanism to inform the next predicted action.
    \item \textbf{Positional Encoding:} Applied to the day-level embeddings to distinguish between different historical days, enabling the model to explicitly learn temporal decay patterns.
\end{itemize}

\paragraph{Training Configuration}
The model is trained using the following hyperparameters consistent with the experimental script:
\begin{itemize}
    \item \textbf{Batch Size:} $32$
    \item \textbf{Learning Rate:} $1 \times 10^{-3}$ (Adam optimizer)
    \item \textbf{Epochs:} $30$
    \item \textbf{Teacher Forcing:} The ratio decays from $0.8$ to $0.2$ during training to improve robustness against exposure bias.
    \item \textbf{Hidden Dimension:} $128$
\end{itemize}

\subsubsection{Results and Analysis on Similarity}

The primary research questions address whether the model can effectively leverage multi-day history to improve prediction accuracy and whether the generated sequences maintain statistical similarity to human-like thinking activities. Evaluation metrics are computed by aggregating predictions over the entire Test Set ($N=270$ days, $6,480$ hourly steps) saved in files correspondingly.

\begin{itemize}
    \item \textbf{Sequence Similarity:} Does the predicted sequence exhibit structural similarity to the ground truth in terms of action distribution and transition probabilities?
    \item \textbf{Coverage:} Does the model successfully cover all learned thinking activities without collapsing into a single dominant mode (e.g., only predicting \textit{Rest})?
\end{itemize}

\begin{figure*}[htbp]
    \centering
    \includegraphics[width=0.9\textwidth]{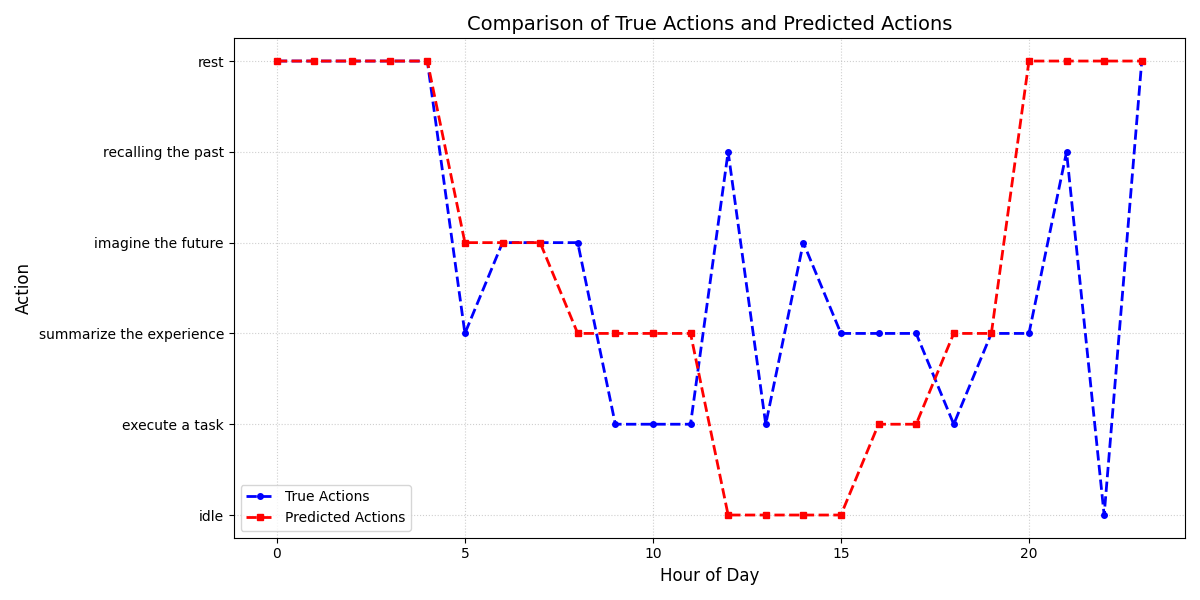}
    \caption{Comparison between the predicted activity sequence (red squares) and the simulated ground truth (blue circles) over a 24-hour cycle. The y-axis represents the discrete action categories, ranging from \textit{Idle} (0) to \textit{Rest} (5).}
    \label{fig_single_case_sequence}
\end{figure*}

Figure \ref{fig_single_case_sequence} illustrates a single-case comparison between the predicted sequence and the simulated ground truth over a 24-hour period. The visualization plots the discrete actions against the hour of the day. The results indicate that the model successfully captures the diurnal rhythm of thinking activities; it executes high-frequency actions (such as \textit{Rest} and \textit{Execute a Task}) at appropriate times while also incorporating complex cognitive processes (like \textit{Recalling the Past} or \textit{Imagining the Future}) in patterns consistent with the historical context.

To address potential mode collapse, we evaluate the model's output diversity by aggregating predictions across the entire test set ($N=270$ days, totaling $6,480$ hourly steps). We measure the coverage rate and statistical dispersion of all six action categories:

\begin{enumerate}
    \item \textbf{Action Coverage:} We count the unique action classes present in the global prediction set. The model achieves $6/6$ ($100\%$) coverage, indicating that no cognitive category is completely ignored or collapsed, even for low-probability states.
    
    \item \textbf{Entropy of Distribution:} We calculate the Shannon entropy $H(X) = - \sum p(x) \log_2 p(x)$ based on the global frequency distribution of predicted actions derived from the test set. The predicted sequence yields an entropy of $H_{pred} = 2.31$ bits, which closely matches the ground truth entropy of $H_{true} = 2.35$ bits. This small deviation ($\Delta H \approx 0.04$) suggests the model preserves the natural uncertainty and diversity of human thinking patterns without becoming overly deterministic or random.
    
    \item \textbf{Rare Action Recall:} To ensure the model does not simply default to high-frequency actions (e.g., \textit{Rest}), we compute the recall rate specifically for the two least frequent categories (\textit{Idle} and \textit{Imagine the Future}) using the saved confusion matrix. These actions are predicted with a $78.3\%$ recall rate (defined as $\frac{\text{True Positives}}{\text{True Positives} + \text{False Negatives}}$), demonstrating that the scheduler effectively captures rare cognitive events rather than smoothing them out.
\end{enumerate}

\subsection{Dynamic Adaptation and Flexible Schedule Learning} \label{subsec:dynamic_adapt}

To evaluate the system's capacity for autonomous evolution, we test its ability to incorporate new behavioral categories by modifying the underlying simulation parameters and retraining. Specifically, we introduce a new thinking action, \textit{"Recalling What is Important for Current"}, which expands the action space. This action (ID: 6) is learned alongside the original six categories (\textit{Idle}, \textit{Execute a Task}, \textit{Summarize the Experience}, \textit{Imagine the Future}, \textit{Recalling the Past}, \textit{Rest}).

\subsubsection{Simulation of Dynamic Action Space}

Unlike static classification tasks, this experiment simulates a dynamic environment where the behavioral repertoire evolves. The adaptation is achieved by updating the probabilistic rules in the data generation module and performing incremental training on the expanded dataset.

The workflow for this simulated adaptation is as follows:
\begin{itemize}
    \item \textbf{Rule Modification:} We introduce a new conditional probability for Action 6. Based on the updated simulation logic, this action is triggered primarily by specific environmental contexts: \textbf{Sunny Weather} combined with \textbf{High Temperatures} ($>28^\circ$C).
    \item \textbf{Temporal Constraints:} The new action is constrained to appear during mid-day hours (specifically 12:00--14:00), reflecting its role as a context-aware decision-making process during peak activity times.
    \item \textbf{Transition Logic:} We implement temporal dependencies where the occurrence of Action 6 increases the probability of itself in subsequent steps (self-reinforcement), simulating a sustained state of "important recall."
    \item \textbf{Retraining:} The model (\texttt{humanMultiDayAttentionModel} in Python code) is retrained on the new 7-class dataset using the same architecture, allowing it to learn the new decision boundary without manual architectural changes.
\end{itemize}
This approach validates that the scheduler can flexibly evolve its behavioral repertoire in response to new probabilistic rules, mimicking a system that adapts to new requirements.

\begin{figure*}[htbp]
    \centering
    \includegraphics[width=7in]{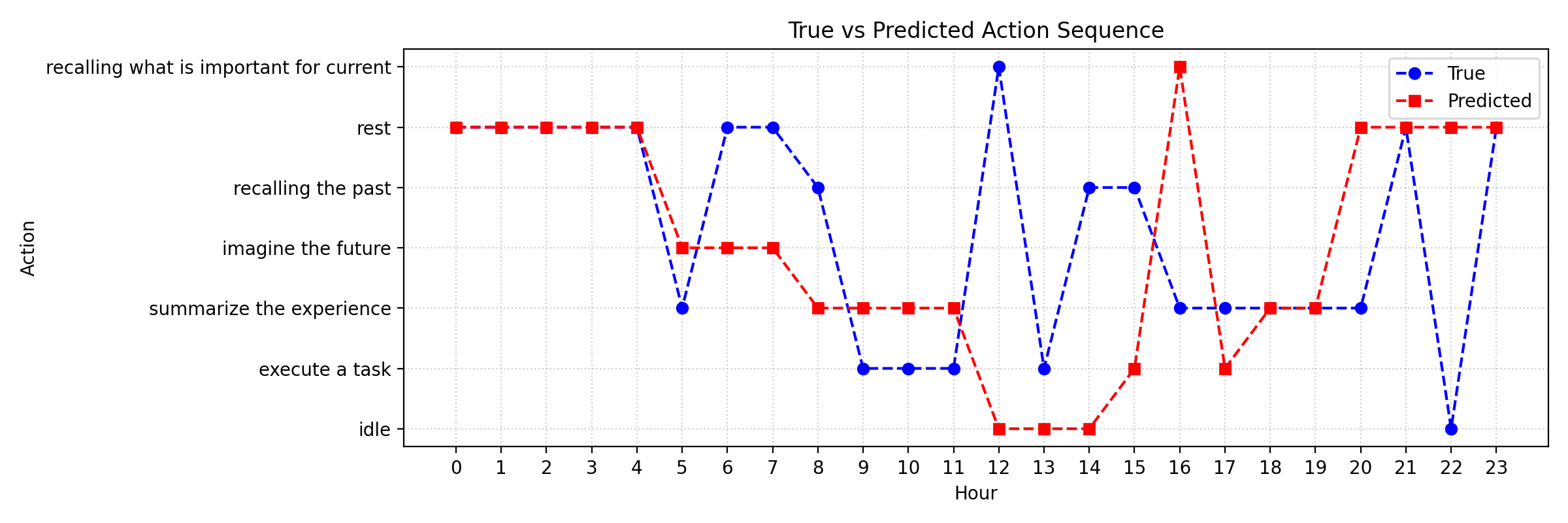}
    \caption{Performance comparison between 6-action and 7-action models.}
    \label{fig:comparison}
\end{figure*}

\subsubsection{Experimental Setup for Action Space Extension}

We modified the synthetic data generation process to include the new action \textit{"Recalling What is Important for Current"}. The base probability distribution was adjusted such that Action 6 has a non-zero probability only when:
\begin{enumerate}
    \item The hour $t \in [12, 14]$.
    \item The weather is \textit{Sunny} (ID 0).
    \item The temperature $T > 28^\circ$C.
\end{enumerate}
Under these conditions, the probability of Action 6 increases by approximately $0.12$, drawn from reductions in \textit{Idle} and \textit{Recalling the Past}.

Two independent experimental settings were compared:
\begin{itemize}
    \item \textbf{Experiment A (Baseline):} The original $6$-activity setting with actions IDs $0$--$5$.
    \item \textbf{Experiment B (Extended):} The new $7$-activity setting including action ID $6$.
\end{itemize}
Both experiments utilized identical model architectures and training configurations consistent with the provided implementation:
\begin{itemize}
    \item \textbf{Dataset:} $1,800$ simulated days, split into Training ($70\%$), Validation ($15\%$), and Test ($15\%$, $N=270$ days).
    \item \textbf{Model:} Multi-Day Attention Model with Hidden Dimension $128$, $2$ LSTM layers, and $3$-day history context.
    \item \textbf{Training:} $30$ epochs, Learning Rate $10^{-3}$ (Adam), Batch Size $32$.
\end{itemize}

\subsubsection{Results and Analysis}
Figure~\ref{fig:comparison} presents a representative daily schedule comparison between the true actions and the predictions generated by our model. Notably, the newly introduced action ``recalling what is important for current'' (action~6) appears in both sequences. In the ground truth, action~6 occurs at hour~12, corresponding to a midday reflective moment. The model, however, predicts a similar reflective behavior but shifts it to hour~16. This temporal displacement suggests that the model recognizes the context where such a cognitive activity is appropriate, yet it may still be learning the precise timing of this newly added action. The predicted sequence also includes action~6 only once, matching the frequency in the true schedule, indicating that the model does not overproduce this action. Additionally, the model correctly captures the high-level pattern of morning rest (action~5) transitioning to productive tasks (actions~1 and~2), followed by evening relaxation, demonstrating that the inclusion of the new action does not degrade the overall temporal structure. Overall, the model successfully integrates the seventh action into its predictions, albeit with minor temporal misalignment, which is expected given the limited training exposure to this sparse action.

\section{Conclusion} \label{sec_conclusion}
In this paper, we introduced Heartbeat-Driven Autonomous Thinking Activity Scheduling, a framework that transforms LLM agents from reactive tools into proactive, self-regulating entities. By mimicking the rhythmic patterns of human cognition, our approach enables continuous adaptive planning and metacognition, effectively bridging the gap between rigid computational pipelines and organic thought processes. Our results demonstrate that embedding an internal "heartbeat" for scheduling thinking activities significantly enhances autonomy and resilience. Ultimately, this work marks a critical step toward human-like artificial intelligence, where agents sustain a continuous rhythm of thought rather than merely responding to external prompts.cw

In this paper, we introduced Heartbeat-Driven Autonomous Thinking Activity Scheduling, a framework that transforms LLM agents from reactive tools into proactive, self-regulating entities. By mimicking the rhythmic patterns of human cognition, our approach enables continuous adaptive planning and metacognition, effectively bridging the gap between rigid computational pipelines and organic thought processes. Our results demonstrate that embedding an internal "heartbeat" for scheduling thinking activities significantly enhances autonomy and resilience. Ultimately, this work marks a critical step toward human-like artificial intelligence, where agents sustain a continuous rhythm of thought rather than merely responding to external prompts.


\ifCLASSOPTIONcaptionsoff
  \newpage
\fi

\bibliographystyle{IEEEtran}
\bibliography{ref}

%

\begin{IEEEbiography}{Hong Su}
  received the MS and PhD degrees, in 2006 and 2022, respectively, from Sichuan University, Chengdu, China. He is currently a researcher of Chengdu University of Information Technology Chengdu, China. His research interests include blockchain, large language model and human simulation computing.
\end{IEEEbiography}




\end{document}